\documentclass[review]{elsarticle}

\usepackage{lineno,hyperref}
\usepackage{booktabs}
\usepackage{threeparttable}
\usepackage{multirow}
\usepackage{tabularx}
\usepackage{amsmath}
\usepackage{amsfonts,amssymb}
\usepackage{xcolor}

\usepackage{amsfonts,amssymb}
\usepackage{xcolor,amsmath}
\usepackage[ruled,vlined]{algorithm2e}
\DontPrintSemicolon 

\SetKwComment{Comment}{\color{green!50!black}//}{}

\SetKwProg{Function}{function}{}{}

\graphicspath{{./figures/}}

\modulolinenumbers[5]

\journal{Elsevier}

\begin{document}

\begin{frontmatter}

\title{Semantic Representation and Dependency Learning \\ for Multi-Label Image Recognition}

\author[sysu]{Tao Pu}
\author[plan]{Mingzhan Sun}
\author[sysu]{Hefeng Wu\corref{correspondingauthor}}
\ead{wuhefeng@gmail.com}
\author[gdut]{Tianshui Chen}
\author[uestc]{Ling Tian}
\author[sysu]{Liang Lin}

\cortext[correspondingauthor]{Corresponding author}



\address[sysu]{Sun Yat-sen University}
\address[plan]{Plan Technology}
\address[gdut]{Guangdong University of Technology}
\address[uestc]{University of Electronic Science and Technology of China}

\begin{abstract}
Recently many multi-label image recognition (MLR) works have made significant progress by introducing pre-trained object detection models to generate lots of proposals or utilizing statistical label co-occurrence enhance the correlation among different categories. However, these works have some limitations: (1) the effectiveness of the network significantly depends on pre-trained object detection models that bring expensive and unaffordable computation; (2) the network performance degrades when there exist occasional co-occurrence objects in images, especially for the rare categories. To address these problems, we propose a novel and effective semantic representation and dependency learning (SRDL) framework to learn category-specific semantic representation for each category and capture semantic dependency among all categories. Specifically, we design a category-specific attentional regions (CAR) module to generate channel/spatial-wise attention matrices to guide model to focus on semantic-aware regions. We also design an object erasing (OE) module to implicitly learn semantic dependency among categories by erasing semantic-aware regions to regularize the network training. Extensive experiments and comparisons on two popular MLR benchmark datasets (i.e., MS-COCO and Pascal VOC 2007) demonstrate the effectiveness of the proposed framework over current state-of-the-art algorithms.
\end{abstract}


\end{frontmatter}


\section{Introduction}
\begin{figure}[!h]
   \centering
   \includegraphics[width=0.80\linewidth]{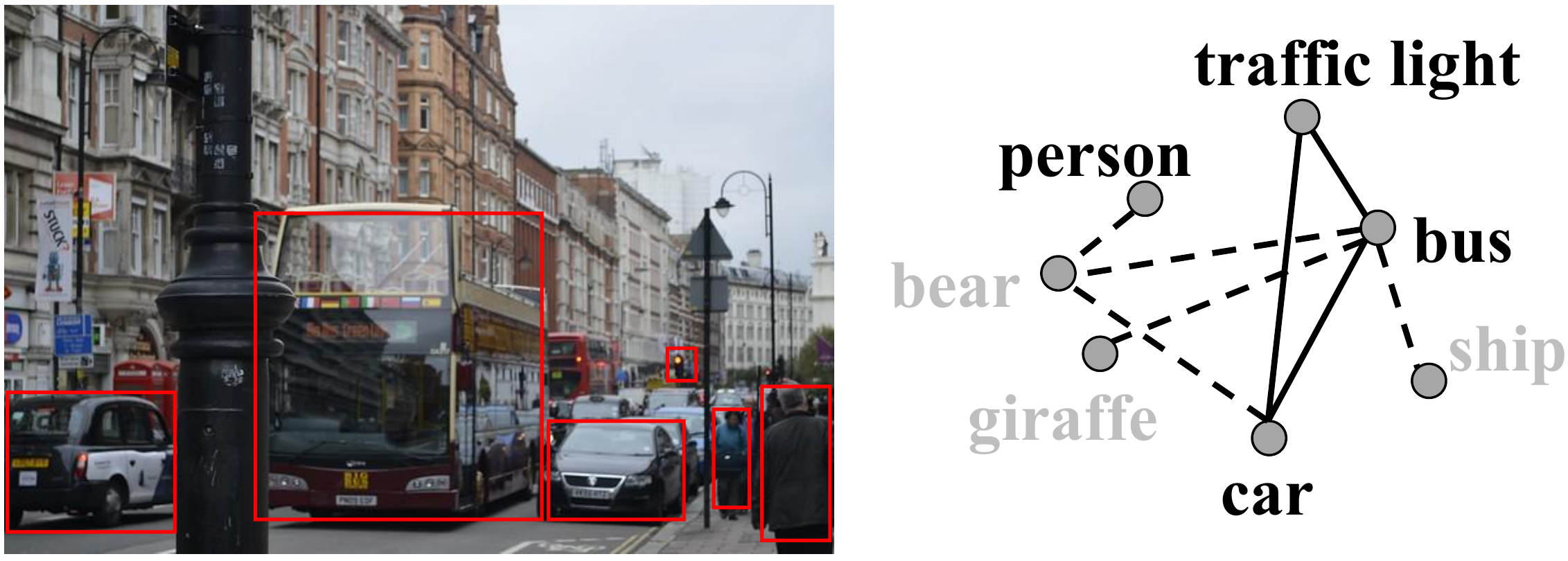}
   \caption{Left: An example of MLR image (object regions are highlighted in red). Right: Corresponding label co-occurrence probability (solid line indicates high co-occurrence probability, and dotted line indicates low co-occurrence probability).}
   \label{fig:motivation}
\end{figure}

Multi-label image recognition (MLR) is a fundamental yet practical task in computer vision, and plays a critical role in wide applications, e.g., human attribute recognition \cite{Wang2017attribute, Li2019pedestrian, chen2021cross}, medical diagnosis \cite{Baltruschat2019comparison, Bustos2020padchest}, content-based image retrieval \cite{Chua1994concept,DengWWHLL18ivc} and recommendation systems \cite{Yang2015pinterest, Tzelepi2018deep}. Compared to its single-label counterpart, there are not only multiple object regions in MLR images but also the semantic dependency among different categories. As shown in Figure \ref{fig:motivation}, localizing each object region makes it easy to extract corresponding representation feature for each category. Meanwhile, it is intuitive that exploiting label co-occurrence among all categories can be helpful to recognize objects with small sizes or occlusions. In particular, the region of ``traffic light" is too small to recognize by its coarse representation feature, but we can utilize the high co-occurrence among ``traffic light", ``car" and ``bus" to infer its existence. Therefore, it is reasonable to take account of focusing semantic-aware regions and exploring semantic dependency among all categories.

Many existing works \cite{Wei2015HCP, Wang2017RDAL, Chen2018RARL, Wang2020IA-GCN, nie2022multi} utilize pre-trained object detection models or object detection mechanisms to generate lots of proposals to coarsely localize object regions and then treat each region as single-label image recognition. However, it not only brings a huge computation cost which is expensive and unaffordable, but also requires sufficient object detection annotations to train object detection models. On the other hand, these works ignore the relationship among different regions which plays a key role in recognizing objects with small sizes or occlusions. Besides introducing object detection mechanism, some works also utilize graph neural network \cite{Chen2019SSGRL, Wu2020AdaHGNN, Ye2020ADD-GCN, ChenCHWLL20aaai, TianZWZZ2022JEST} to directly model label dependency among different categories. However, these methods construct a graph coarsely by counting the label co-occurrence from the training set, which may degrade the model performance when there exist occasional co-occurrence objects in images.

To address above issues, we propose a novel and effective semantic representation and dependency learning (SRDL) framework that consists of two crucial module, i.e., category-specific attentional region (CAR) module and object erasing (OE) module, to learn more discriminative and robust semantic representation for each category and capture semantic dependency among all categories. More specifically, we design the CAR module to guide model to focus on semantic-aware regions by generating channel/spatial-wise attention maps for each category, and we design the OE module to capture semantic dependency among all categories by adaptively erasing obvious semantic-aware regions to regularize network training.

The contributions of this work are summarized into four folds: 1) We propose a novel and effective semantic representation and dependency learning (SRDL) framework to learn more discriminative semantic representation for each category and capture semantic dependency among all categories. 2) We design a category-specific attentional regions (CAR) module to learn category-specific semantic representation feature by generating channel/spatial-wise attention maps to guide model in focusing on semantic-aware regions. 3) We design an object erasing (OE) module to capture semantic dependency among all categories by adaptively erasing obvious semantic-aware regions to regularize network training. 4) We conduct extensive experiments on two MLR benchmark datasets (i.e., MS-COCO and Pascal VOC 2007) to demonstrate the effectiveness of the proposed framework. We also conduct ablative studies to analyze the actual contributions of each module for profound understanding.

\section{Related Work}

Since it is more fundamental yet practical than its single-label counterpart, multi-label image recognition receives increasing attention \cite{Chen2019SSGRL, Chen2019ML-GCN, Zhou2021DSDL, Gao2021MCAR}, and recent works gradually explore this task in partial labels scenario \cite{Durand2019Learning, Huynh2020Interactive, Pu2022SARB, Chen2022SST, Pu2022SARB-journal, Chen2022HST} or few-shot scenario \cite{Narayan2021discriminative, Ben2021semantic, ChenLCHW22pami}. To solve this task, lots of efforts are dedicated to localize discriminative regions by object detection models or attention mechanism. Another line of works propose to introduce label co-occurrence to capture label dependency among all categories to regularize the network training. Besides, there are also general multi-class image recognition methods that use different approaches. In this section, we briefly review the related approaches by roughly categorizing them into four main directions as following.

\noindent{\textbf{Object detection based methods. }} With the rapid development of object detection \cite{Redmon2016YOLO, Ren2017FasterRCNN}, utilizing pre-trained object detection models to localize multiple object regions and then recognize each region with CNNs is feasible. HCP \cite{Wei2015HCP} generates lot of proposals using objection detection models and treats each proposal as a single-label image recognition problem. IA-GCN \cite{Wang2020IA-GCN} involves a global branch modeling the whole image and a region-based branch exploring dependencies among ROIs. MACR \cite{Zhao2020MCAR} propose an unsupervised multi-object cross-domain detection method which improve the detection performance by enforcing consistent object category predictions between the multi-label prediction over global features and the object classification over detection region proposals. However, directly exploiting object detection models bring some disadvantages: 1) generating lots of proposals brings a huge computation cost; 2) it makes the network performance depend on pre-trained object detection models which need time-consuming annotating for exhaustive object bounding boxes. 

\noindent{\textbf{Attention mechanism based methods. }} As above mentioned, previous object detection based methods not only have high inference time, but also require lots of bounding box annotations for pre-training object detection models. Therefore, recent works exploit attention mechanism to capture attentional regions for each category. For instance, RADL \cite{Wang2017RDAL} introduces a spatial transformer label to locate attentional regions from feature maps in a region-proposal-free way, RARL \cite{Chen2018RARL} proposes a recurrent attention reinforcement learning framework to iteratively discover a sequence of attentional and informative regions and MCAR \cite{Gao2021MCAR} designs a parameter-free region localization module to decompose an image into multiple object regions. Besides, SSGRL \cite{Chen2019SSGRL} proposes a semantic decoupling (SD) module that incorporates category semantics to guide capturing attentional regions for learning semantic-specific representation. ADD-GCN\cite{Ye2020ADD-GCN} proposes a semantic attention module (SAM) for locating semantic regions and producing content-aware category representations for each image. Although these works get rid of the limitation of object detection models, they merely focus on spatial-wise attention while ignoring the channel-wise attention.

\noindent{\textbf{Label co-occurrence based methods. }} In multi-label image recognition, the label dependency among different categories is crucial for recognizing each category, especially for the complex variations in viewpoint, scale, illumination, occlusion. To exploit the label dependency, Order-free RNN \cite{Chen2018Order-FreeRNN} allows the network to observe proper label sequences using a confidence-ranked LSTM rather than requires predetermined label orders for training. Similarly, PLA \cite{Yazici2020PLA} proposes two alternative losses which dynamically order the labels based on the prediction of the LSTM model. With the in-depth study of graph neural network, more and more works adopt graph neural network to explore the label dependency \cite{Chen2019SSGRL, Chen2019ML-GCN, Chen2021P-GCN}. SSGRL \cite{Chen2019SSGRL} constructs a gated graph neural network by counting the label co-occurrence possibilities to model the label dependency among all categories. ML-GCN \cite{Chen2019ML-GCN} proposes a GCN based model to learn inter-dependent object classifiers from prior label representations and P-GCN \cite{Chen2021P-GCN} proposes prediction learning GCN to encode such features into inter-dependent image-level prediction scores. However, although these works achieve remarkable progress by directly introducing statistical label co-occurrence to model semantic dependency among different categories, it may easily degrade their model performance when there exist occasional co-occurrence objects in images, especially for the rare categories.

\noindent{\textbf{General MLR methods. }} Besides the above-mentioned methods, recently more and more works have gradually improved MLR performance from some general perspectives, e.g., loss function, network architecture, etc. For instance, considering the positive-negative imbalance dominates the optimization process of MLR, ASL \cite{Ben2020ASL} proposes a novel asymmetric loss (ASL) which operates differently on positive and negative samples. Different from previous works, CSRA \cite{Zhu2021CSRA} proposes a class-specific residual attention (CSRA) module to generate class-specific features for every category by proposing a simple spatial attention score, and then combines it with the class-agnostic average pooling feature. As Transformer has achieved great success
in natural language processing (NLP) tasks \cite{Devlin2018Bert}, many works adopt Transformer for addressing multi-label image recognition, e.g., C-Trans \cite{Lanchantin2021C-Trans} proposes a general classification framework that leverages Transformers to exploit the complex dependencies among visual features and labels, TDRG \cite{Zhao2021TDRG} introduces a Transformer-based dual relation framework that constructs complementary relationships by exploring two aspects of correlation, and M3TR \cite{Zhao2021M3TR} proposes the multi-modal multi-label recognition Transformers with ternary relationship learning for inter-and-intra modalities.

Different from these above methods, the proposed framework introduces two novel and effective modules, in which the CAR module generates channel/spatial-wise attention map to guide model in focusing on semantic-aware regions to learn more discriminative category-specific semantic representation and the OE module adaptively erases semantic-aware regions to guide model in indirectly capturing semantic dependency among all categories to learn more contextualized category-specific semantic representation. 

\section{Proposed Framework}

\subsection{Overview}
\begin{figure}[!hb]
   \centering
   \includegraphics[width=0.98\linewidth]{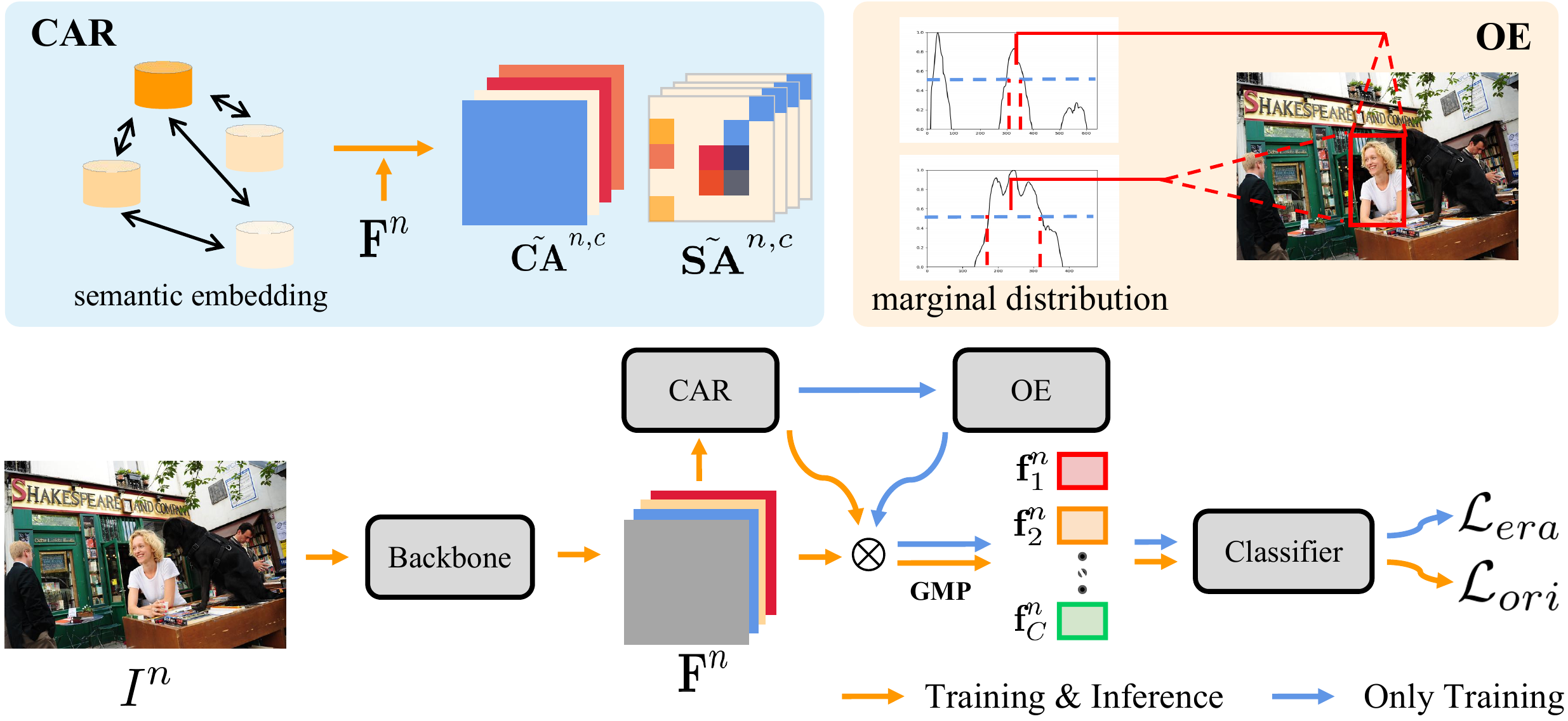}
   \caption{An overall illustration of our Semantic Representation and Dependency Learning framework. Given an input image, we first feed it into a CNN backbone to extract feature map. Then, the category-specific attentional regions (CAR) module incorporates contextualized semantic feature to generate channel/spatial-wise attention map to learn category-specific semantic representation, and the object erasing (OE) module decomposes the spatial-wise attention map into two marginal distributions to adaptively erase obvious semantic-aware region  for regularizing network training. During inference, the OE module is removed.}
   \label{fig:framework}
\end{figure}

In this section, we introduce the proposed SRDL framework which consists of two crucial modules, i.e., category-specific attentional regions (CAR) module and object erasing (OE) module, to learn more discriminative semantic representation for each category and capture semantic dependency among different categories. In particular, the CAR module guides model to focus on semantic-aware regions by incorporating contextualized semantic features to generate channel/spatial-wise attention maps for each category to learn more discriminative category-specific semantic representation. And the OE module regularizes network training by decomposing the spatial-wise attention map to adaptively erase obvious semantic-aware region to capture semantic dependency among all categories. Figure \ref{fig:framework} illustrates an overall pipeline of the proposed framework.

Specifically, given an input image $I^n$, we first feed it into a fully convolutional network to extract its feature map $\textbf{F}^n \in \mathbb{R}^{w \times h \times d}$ where $w$, $h$, and $d$ are the width, height and channel number of the feature map, formulated as
\begin{equation}
    \textbf{F}^n = f_{cnn}(I^n).
\end{equation}

Then, we adopt a graph neural network to learn contextualized semantic features, and utilize the CAR module to incorporate these contextualized semantic features to generate a channel-wise attention vector $\tilde{\textbf{CA}}^{n,c} \in [0,1]^{w \times h \times d}$ and a spatial-wise attention map $\tilde{\textbf{SA}}^{n,c} \in [0,1]^{w \times h \times d}$ for each category $c$. By fusing these attention maps and original feature map, we obtain weighted feature map $\textbf{F}^{n,a}_c$, and semantic representation $\textbf{f}^n_c$ for each category $c$. Finally, the network utilizes a linear classifier followed by a sigmoid function to compute the probability score vectors:
\begin{equation}
  [p^n_1, p^n_2, \cdots, p^n_C]=\psi(\phi([\textbf{f}^n_1, \textbf{f}^n_2, \cdots, \textbf{f}^n_C])),
\end{equation}
where $\psi$ is the sigmoid function and $\phi$ is a linear classifier. We utilize these probability score vectors and the corresponding label vector $\textbf{y}^n=\{y^n_1, y^n_2, \cdots, y^n_C\}$ to compute loss. 

Besides, during training phase, we select some categories based on predicted probability scores, and feed these corresponding spatial-wise attention maps into the OE module. Then, the OE module adaptively erases the obvious semantic-aware regions and generate corresponding representation feature $\hat{\textbf{f}}^n_c$. The proposed framework utilizes these features to regularize network training for capturing semantic dependency among different categories.

\subsection{Discovering Semantic-aware Regions}
Different from previous works, more and more recent works propose to introduce category semantic embedding to guide learning category-specific semantic representation \cite{Chen2019SSGRL, Wu2020AdaHGNN, Ye2020ADD-GCN}. However, most existing works only explore the spatial-wise attention message, ignoring the channel-wise attention message that also plays an important role in learning category-specific semantic representation feature. Therefore, we design a category-specific attentional regions (CAR) module to generate channel/spatial-wise attention maps to guide model to focus on semantic-aware regions.

\begin{figure}[!h]
   \centering
   \includegraphics[width=0.98\linewidth]{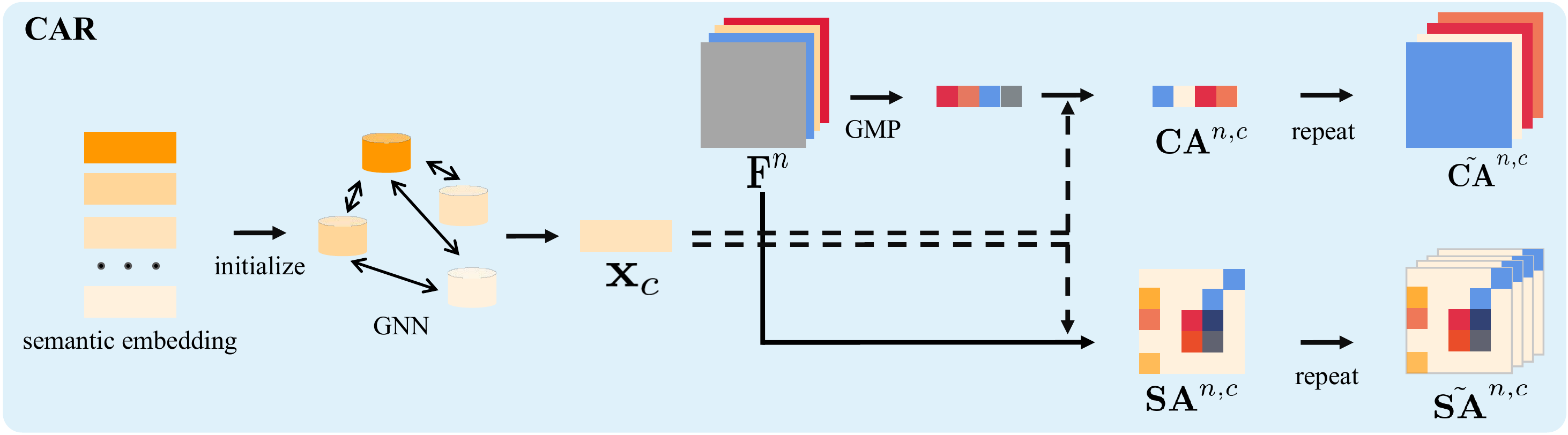}
   \caption{Illustration of category-specific attentional regions (CAR) module. First, we compute the contextualized semantic embeddings by exploiting a graph $\mathcal{G}$. And then, we incorporate the contextualized semantic embedding to generate channel/spatial-wise attention maps.}
   \label{fig:CAR}
\end{figure}

Like previous works \cite{Chen2019SSGRL, Chen2019ML-GCN, Ye2020ADD-GCN}, we also adopt semantic embedding to guide model learning category-specific representation. However, similar to ML-GCN \cite{Chen2019ML-GCN}, we consider that contextualized semantic embedding is better than word embedding for stable training. Thus, we introduce the graph $\mathcal{G}=\{V, A\}$, in which this graph takes the node features $A^l \in \mathbb{R}^{C \times d'}$ and adjacency matrix $V \in \mathbb{R}^{C \times C}$ as inputs (where $C$ demotes the number of categories and $d'$ demotes the dimension of node features), and updates the node features as $A^{l+1} \in \mathbb{R}^{C \times d'}$, formulated as
\begin{equation}
    A^{l+1} = h( \hat{V} A^l W^l ),
\end{equation}
where $W^l \in \mathbb{R}^{d' \times d'}$ is a transformation matrix to be learned and $\hat{V} \in \mathbb{R}^{C \times C}$ is the normalized version of matrix $V$, and $h(\cdot)$ denotes a non-linear operation, which is acted by LeaklyReLU \cite{Maas2013RectifierNI} in our experiments. In addition, we randomly initialize adjacency matrix $V$ and initialize nodes features $A$ by pre-trained GloVe word embedding \cite{Pennington2014GloVe}. At the last layer $L$, we extract the corresponding row in $A^L$ for each category $c$ as the  contextualized semantic embedding $\textbf{x}_c \in \mathcal{R}^{d'}$.

Then, we incorporate the contextualized semantic embedding $\textbf{x}_c$ to generate channel/spatial-wise attention vector/matrix for each category $c$:
\begin{gather}
  \textbf{CA}^{n,c} = \psi(f_c(tanh(f_{ca}(GMP(\textbf{F}^{n})) \odot f_w(\textbf{x}_c)))), \\
  \textbf{SA}^{n,c}_{x,y} = \psi(f_s(tanh(f_{sa}(\textbf{F}^{n}_{x,y}) \odot f_w(\textbf{x}_c)))),
\end{gather}
where $\psi(\cdot)$ is the sigmoid function, $tanh(\cdot)$ is the hyperbolic tangent function, $\odot$ is the element-wise multiplication operation, $GMP(\cdot)$ is the global max pooling operation, $f_c(\cdot)$, $f_s(\cdot)$, $f_{ca}(\cdot)$, $f_{sa}(\cdot)$ and $f_w(\cdot)$ are fully connected layers.

Thereby we obtain a channel-wise attention coefficient vector $\textbf{CA}^{n,c} \in [0,1]^{d}$ and a spatial-wise attention coefficient matrix $\textbf{SA}^{n,c} \in [0,1]^{w \times h}$. To align the dimension with the feature map $\textbf{F}^n$, we perform repeating on these attention maps, resulting channel-wise attention map $\tilde{\textbf{CA}}^{n,c} \in [0,1]^{w \times h \times d}$ and spatial-wise attention map $\tilde{\textbf{SA}}^{n,c} \in [0,1]^{w \times h \times d}$. Using these channel/spatial-wise attention maps, the CAR module guides network to focus on semantic-aware regions for each category $c$:
\begin{equation}
    \mathbf{\hat{F}}^n_c = 0.5 * \textbf{F}^n \odot \tilde{\textbf{SA}}^{n,c} + 0.5 * \textbf{F}^n \odot \tilde{\textbf{CA}}^{n,c}.
\end{equation}

Although most state-of-the-art CNNs utilize global average pooling (GAP) to encode feature maps to a feature vector, previous MLR work \cite{Chen2019ML-GCN} found that replacing GAP with global max pooling (GMP) leads to improved performance, so we perform GMP on $\mathbf{\hat{F}}^n_c$ to compute category-specific feature representation $\textbf{f}^{n}_c \in \mathbb{R}^{d}$.

\subsection{Learning Semantic Dependency}
Different from current works that directly utilize the statistical label co-occurrence \cite{Chen2019ML-GCN, Chen2019SSGRL}, we argue that it may degrade model performance when there exist occasional co-occurrence objects in test images, especially for rare categories. Specifically, we design a novel and effective object erasing (OE) module that adaptively erases the obvious semantic-aware regions during the training stage to guide model to learn more robust representation for objects with small sizes or occlusions and indirectly capture semantic dependency among all categories.

\begin{figure}[!h]
   \centering
   \includegraphics[width=0.98\linewidth]{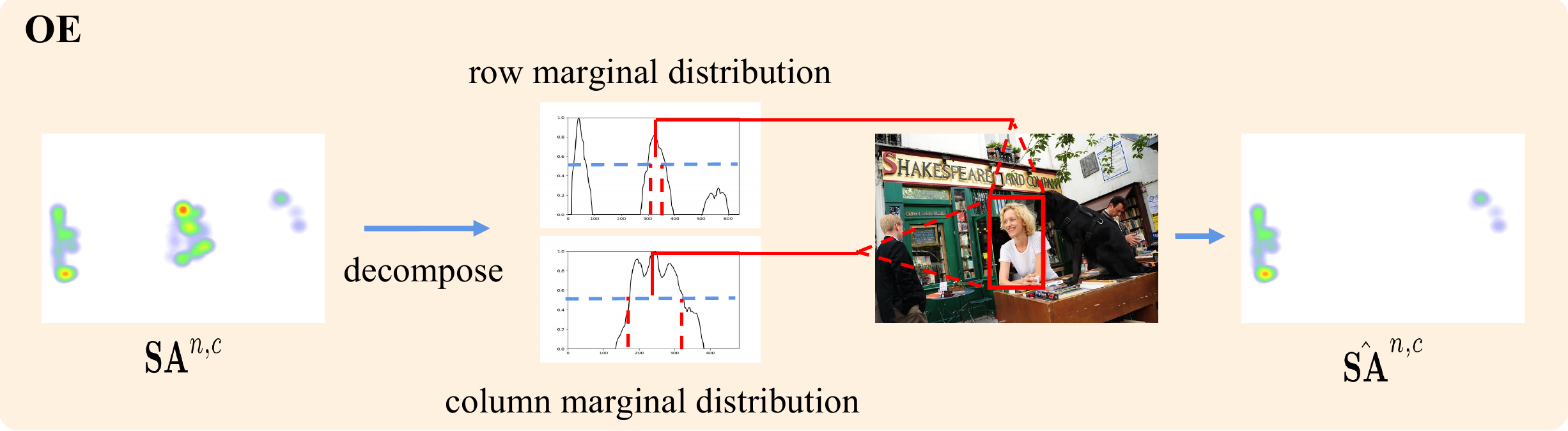}
   \caption{Illustration of object erasing (OE) module. Given the spatial-wise attention map, we firstly decompose it into two marginal distributions along row and column. Then, we utilize a constant threshold $\alpha$ to localize the obvious semantic-aware region, and erase corresponding attention coefficient to generate erased spatial-wise attention map.}
   \label{fig:OE}
\end{figure}

In the OE module, we select the most likely semantic-aware region to erase based on the corresponding spatial attention matrix $\textbf{SA}^{n,c}$. One of the simplest schemes is erasing the constant-sized area around the highest attention coefficient in attention map. However, it is ineffective because the scale of different object regions in each category considerably varies. Moreover, it is impractical to design different erased region sizes for different categories for different datasets.
To adaptively localize semantic-aware regions, we follow \cite{Gao2021MCAR} to decompose the category-specific spatial-wise attention matrix $\textbf{SA}^{n,c}$ into two marginal distributions along row and column $\textbf{m}^{n,c}_{x} \in [0,1]^w, \textbf{m}^{n,c}_{y} \in [0,1]^h$, formulated as
\begin{gather}
    \textbf{m}^{n,c}_{x} = Norm(\max_{1 \le y \le h}{\textbf{SA}^{n,c}_{x,y}}), \\
    \textbf{m}^{n,c}_{y} = Norm(\max_{1 \le x \le w}{\textbf{SA}^{n,c}_{x,y}})
\end{gather}
where $Norm(\cdot)$ is a min-max normalization function. Then we localize object regions based on these two marginal distributions:
\begin{gather}
    m^{n,c}_{x,i} \ge \alpha, \quad s.t. \quad i={1,2,...,w} \\
    m^{n,c}_{y,j} \ge \alpha, \quad s.t. \quad j={1,2,...,h}
\end{gather}
where $\alpha \in (0,1)$ is a constant threshold. However, it results in too many regions being erased when multiple objects exist in the input image. Therefore, we select the most obvious object region to erase, which means we pick the interval contained in the global maximum peak for the case of multiple local maximum peaks and choose the widest interval for multiple global maximum peaks.

Considering erasing semantic-aware regions on all categories may disturb the training process, which results in slow convergence and the performance degradation, we sort the probability score vectors following a descending order and select the \textit{top K} category spatial-wise attention matrices (denote as $\{\textbf{SA}^{i,c}\}^{\textit{topK}}_{i=1}$), because the high category confidence means that the corresponding object presents in the image with a high probability.

\subsection{Optimization}
Given a dataset that contains $N$ training samples $\{I^n, \textbf{y}^n\}^{N}_{n=1}$, in which $I^n$ is the $n$-th image and $\textbf{y}^n=\{y^n_1, y^n_2, \cdots, y^n_C\}$ is the corresponding label vector. $y^n_c$ is assigned as 1 if the $n$-th image contains category $c$ and 0 otherwise. In training, given an image $I^n$, we obtain a probability score vector $\textbf{p}^n=[p^n_1, p^n_2, \cdots, p^n_C]$ from the original feature map and a probability score vector $\hat{\textbf{p}}^n=[\hat{p}^n_1, \hat{p}^n_2, \cdots, \hat{p}^n_C]$ from the erased feature map.

Following previous works, we adopt the binary cross entropy as the objective loss function:
\begin{equation}
    \mathcal{L}_{ori} = \sum^N_{n=1}\sum^C_{c=1}{y^n_c log p^n_c + (1-y^n_c) log (1-p^n_c)},
\end{equation}
\begin{equation}
    \mathcal{L}_{era} = \sum^N_{n=1}\sum^C_{c=1}{y^n_c log \hat{p}^n_c + (1-y^n_c) log (1-\hat{p}^n_c)}.
\end{equation}
Therefore, the final loss is defined as summing the above two losses over all samples, formulated as
\begin{equation}
    \mathcal{L} = \mathcal{L}_{ori} + \mathcal{L}_{era}.
\end{equation}

\section{Experiments}

\subsection{Experimental Settings}

\noindent{\textbf{Dataset. }} Following previous MLR works, we conduct extensive experiments on MS-COCO \cite{Lin2014COCO} and Pascal VOC 2007 \cite{Everingham2010Pascal} datasets for fair evaluation. As a challenging and practical MLR benchmark dataset, MS-COCO covers 80 daily categories and contains about 120k images which is further divided into a training set of 82,081 images and a validation set of 40,137 images. Pascal VOC 2007 is one of the most widely used MLR benchmark datasets, and most of existing works reported their results on this dataset. It contains a trainval set of 5,011 images and a test set of 4,952 and covers 20 daily categories.

\noindent{\textbf{Evaluation Metric. }} For a fair comparison, we adopt the mean average precision (mAP) over all categories for evaluation. Moreover, we also follow most current works to adopt the overall precision (OP), recall (OR), F1-measure (OF1) and per-class precision (CP), recall (CR), F1-measure (CF1), which are defined as below
\begin{gather}
 OP=\frac{\sum_{i}{N^c_i}}{\sum_{i}{N^p_i}}, \quad CP=\frac{1}{C} \sum_{i}{ \frac{N^c_i}{N^p_i} }, \\
 OR=\frac{\sum_{i}{N^c_i}}{\sum_{i}{N^g_i}}, \quad CR=\frac{1}{C} \sum_{i}{ \frac{N^c_i}{N^g_i} }, \\
 OF1=\frac{2 \times OP \times OR }{OP+OR},  \quad CF1=\frac{2 \times CP \times CR }{CP+CR},
\end{gather}
where $N^c_i$ is the number of images that are correctly predicted for the $i$-th category, $N^p_i$ is the number of predicted images for the $i$-th category, $N^g_i$ is the number of ground truth images for the $i$-th category. Among these metrics, mAP, CF1, OF1 are the most important metrics that can provide more comprehensive evaluation.

\subsection{Implementation Details}

To fairly compare with state of the arts, we adopt ResNet-101 \cite{He2016ResNet} as the backbone to extract feature maps. Following previous works \cite{Chen2019SSGRL, Wu2020AdaHGNN, Chen2021P-GCN}, we initialize the parameters of the ResNet-101 with those pre-trained on the ImageNet \cite{Deng2009Imagenet} dataset. $f_{ca}(\cdot)$ and $f_{sa}(\cdot)$ are implemented by a 2048-to-2048 fully-connected layer, $f_w(\cdot)$ is implemented by a 300-to-2048 fully-connected layer, $f_c(\cdot)$ is implemented by a 2048-to-2048 fully connected layer and $f_s(\cdot)$ is implemented by a 2048-to-1 fully connected layer.

In training for a total of 20 epochs, we use the ADAM algorithm \cite{Kingma2015Adam} with a batch size of 16, momentums of 0.999 and 0.9, and a weight decay of $5 \times 10^{-4}$ and set the original learning rate to $10^{-5}$ and divide it by 10 for every 10 epochs. During training, the input image is resized to 512 $\times$ 512, and we randomly choose a number from \{512, 448, 384, 320, 256\} as the width and height to crop patch. Finally, the cropped patch is further resized to 448 $\times$ 448. Then we randomly perform horizontal flip and perform normalization. During inference, the OE module is removed and the image is resized to 448 $\times$ 448 for evaluation. All experiments are implemented by PyTorch \cite{Paszke2019Pytorch}.

\begin{table}[!h]
  \centering
  \footnotesize
  \begin{tabular}{c|c|c|c|c}
  \hline
  Method & Publication & Input Size & Backbone & mAP (\%) \\
  \hline
  \hline
  \centering CNN-RNN \cite{Wang2016CNN-RNN} & CVPR 2016 & - & VGG16 & 61.2  \\
  \centering SRN \cite{Lee2018ML-ZSL} & CVPR 2018 & 224 $\times$ 224 & ResNet-101 & 77.1 \\
  \centering ACfs \cite{Guo2019ACfs} & CVPR 2019 & 228 $\times$ 228 & ResNet-101 & 77.5 \\
  \hline
  \hline
  \centering ResNet-101 &  & 448 $\times$ 448 & ResNet-101 & 77.1 \\
  \centering SSGRL* \cite{Chen2019SSGRL} & ICCV 2019 & 448 $\times$ 448 & ResNet-101 & 81.9 \\
  \centering ML-GCN* \cite{Chen2019ML-GCN} & CVPR 2019 & 448 $\times$ 448 & ResNet-101 & 80.9 \\
  \centering DSDL \cite{Zhou2021DSDL} & AAAI 2021 & 448 $\times$ 448 & ResNet-101 & 81.7 \\
  \centering MACR \cite{Gao2021MCAR} & TIP 2021 & 448 $\times$ 448 & ResNet-101 & 81.9 \\
  \centering P-GCN \cite{Chen2021P-GCN} & TPAMI 2021 & 448 $\times$ 448 & ResNet-101 & \textbf{83.2} \\
  \hline
  \hline
  \centering Ours &  & 448 $\times$ 448 & ResNet-101 & 82.9 \\
  \hline
  \end{tabular}
  \vspace{2pt}
  \caption{Comparison of our framework and current state-of-the-art competitors for multi-label image recognition on the MS-COCO dataset. * indicates the result is reproduced by using the open-source code, and - denotes the corresponding result is not provided.}
  \label{tab:coco-results}
\end{table}

\begin{table}[!h]
 \centering
 \footnotesize
 \begin{tabular}{c|c|c|c|c|c|c|c}
 \hline
 Method & Publication  & CP & CR & CF1 & OP & OR & OF1 \\
 \hline
 \hline
 \centering SRN \cite{Lee2018ML-ZSL}  & CVPR 2018  & 81.6 & 65.4 & 71.2 & 82.7 & 69.9 & 75.8 \\
 \centering ACfs \cite{Guo2019ACfs} &  CVPR 2019  & 77.4 & 68.3 & 72.2 & 79.8 & 73.1 & 76.3 \\
 \hline
 \hline
 \centering ResNet-101 &    & 72.7 & {72.3} & 72.5 & 77.4 & {75.5} & 76.5 \\
 \centering SSGRL* \cite{Chen2019SSGRL} &  ICCV 2019  & 84.2 & 70.3 & 76.6 & 85.8 & 72.4 & 78.6 \\
 \centering ML-GCN* \cite{Chen2019ML-GCN}  & CVPR 2019  & 82.5 & 69.2 & 75.3 & 86.0 & 72.6 & 78.7 \\
 \centering DSDL \cite{Zhou2021DSDL}  & AAAI 2021  & 84.1 & 70.4 & 76.7 & 85.1 & 73.9 & 79.1 \\
 \centering P-GCN \cite{Chen2021P-GCN} & TPAMI 2021  & 84.9 & \textbf{72.7} & \textbf{78.3} & 85.0 & \textbf{76.4} & \textbf{80.5} \\
 \hline
 \hline
 \centering Ours &   & \textbf{85.4} & 70.8 & {77.4} & \textbf{87.1} & 74.8 & \textbf{80.5}\\
 \hline
 \end{tabular}
 \vspace{2pt}
 \caption{Comparison of our framework and current state-of-the-art competitors for multi-label image recognition on the MS-COCO dataset in six other metrics (\%).}
 \label{tab:coco-results-2}
\end{table}

\begin{table}[!h]
  \centering
  \footnotesize
  \begin{tabular}{c|c|c|c}
  \hline
  Method & Publication & Input Size & Inference Time \\
  \hline
  \hline
  \centering ResNet-101 & & 448 $\times$ 448 & 31.54 ms \\
  \centering SSGRL & ICCV 2021 & 448 $\times$ 448 & 33.36 ms \\
  \centering ML-GCN & CVPR 2021 & 448 $\times$ 448 & 32.40 ms \\
  \centering P-GCN & TPAMI 2021 & 448 $\times$ 448 & 32.33 ms \\
  \centering MACR (Global) & TIP 2021 & 448 $\times$ 448 & 32.48 ms \\
  \centering MACR (Global-local) & TIP 2021 & 448 $\times$ 448 & 123.47 ms \\
  \hline
  \hline
  \centering Ours & & 448 $\times$ 448 & 32.58 ms \\
  \hline
  \end{tabular}
  \vspace{2pt}
  \caption{Inference time comparison of our framework and current state-of-the-art competitors for multi-label image recognition on the MS-COCO dataset.}
  \label{tab:inference-time}
\end{table}

\subsection{Comparison with State-of-the-art algorithms}
To show the effectiveness of our SRDL framework, we compare it with the following state-of-the-art algorithms: SSGRL \cite{Chen2019SSGRL}, ML-GCN \cite{Chen2019ML-GCN}, DSDL\cite{Zhou2021DSDL}, MCAR \cite{Gao2021MCAR}, P-GCN\cite{Chen2021P-GCN}, GCN-MS-SGA \cite{Liang2022GCN-MS-SGA}. However, considering the input size inconsistence among different works (e.g., $224 \times 224$, $384 \times 384$, $448 \times 448$, $576 \times 576$) greatly hinder the fair comparison \cite{Zhao2021M3TR, Liu2021Query2label}, we set a same input size ($448 \times 448$) to train and evaluate models for fair comparison.

Besides, MCAR \cite{Gao2021MCAR} is a two-stream algorithm that first utilizes its global stream to coarsely predict and then exploits its local stream to precisely predict based on previous predictions. And the local stream of MCAR \cite{Gao2021MCAR} brings a huge complexity which results in costing more inference time, i.e., the average inference time per image of merely utilizing global stream is 18.8 ms while the average inference time per image of utilizing global and local stream is 82.8 ms. Thus, we compare the proposed SRDL framework with the global stream of MCAR \cite{Gao2021MCAR}.

\noindent{\textbf{Performance on MS-COCO. }} We first present the comparison results on the MS-COCO dataset as shown in Table \ref{tab:coco-results}. We find that our proposed framework outperforms the traditional multi-label recognition methods (e.g., SSGRL, ML-GCN, DSDL) by a considerable margin. Specifically, our framework obtains a second best mAP of 82.9\%, slightly lower than that of the best performing P-GCN by 0.3\%. Besides, both using marginal distributions of feature maps to localize attentional regions, our proposed framework obviously outperforms MACR, which demonstrates the effectiveness of learning semantic dependency among categories. Table \ref{tab:coco-results-2} shows more comparison results using other metrics on MS-COCO. It can be observed that our framework achieves very comparable results with the best-performing method P-GCN and performs better in the CP and OP metrics. It is worth noting that our framework does not need any pre-trained object detection model to generate lots of proposals to localize object regions, or any statistical label co-occurrence information to capture semantic dependency. This further shows that the effectiveness of the proposed framework for general multi-label image recognition.

Besides the above performance comparison, we also present the inference time comparison on the MS-COCO dataset, as shown in Table \ref{tab:inference-time}. For fair comparison, we measure the inference time of each method using the same input size of 448 $\times$ 448 in milliseconds (ms) on GeForce RTX 3070 GPU. As shown, most multi-label image recognition methods have similar complexity due to the simple operations in the framework and the acceleration of cuDNN and PyTorch. It is worth noting that the total time of MCAR (Global-local) is about 4 times that of the baseline ResNet-101 because it requires repeated inference for better accuracy. 

\begin{table}[!h]
  \centering
  \small
  \begin{tabular}{c|c|c|c|c}
  \hline
  Method & Publication & Input Size & Backbone & mAP (\%) \\
  \hline
  \hline
  CNN-RNN \cite{Wang2016CNN-RNN} & CVPR 2016 & - & VGG16 & 84.0 \\
  HCP \cite{Wei2015HCP} & TPAMI 2015 & - & VGG16 & 90.9 \\
  RDAL \cite{Wang2017RDAL} & CVPR 2017 & 448 $\times$ 448 & VGG16 & 91.3 \\
  RARL \cite{Chen2018RARL} & AAAI 2018 & 448 $\times$ 448 & VGG16 & 91.3 \\
  \hline
  \hline
  \centering ResNet-101 &  & 448 $\times$ 448 & ResNet-101 & 91.8 \\
  \centering SSGRL* \cite{Chen2019SSGRL} & ICCV 2019 & 448 $\times$ 448 & ResNet-101 & 92.8 \\
  \centering ML-GCN* \cite{Chen2019ML-GCN} & CVPR 2019 & 448 $\times$ 448 & ResNet-101 & 92.0 \\
  \centering DSDL \cite{Zhou2021DSDL} & AAAI 2021 & 448 $\times$ 448 & ResNet-101 & 94.4 \\
  \centering MACR \cite{Gao2021MCAR} & TIP 2021 & 448 $\times$ 448 & ResNet-101 & 93.4 \\
  \centering P-GCN \cite{Chen2021P-GCN} & TPAMI 2021 & 448 $\times$ 448 & ResNet-101 & 94.3 \\
  \centering GCN-MS-SGA \cite{Liang2022GCN-MS-SGA} & NEUCOM 2022 & 448 $\times$ 448 & ResNet-101 & 94.2 \\
  \hline
  \hline
  \centering Ours & & 448 $\times$ 448 & ResNet-101 & \textbf{94.6} \\
  \hline
  \end{tabular}
  \vspace{2pt}
  \caption{Comparison of our framework and current state-of-the-art competitors for multi-label image recognition on the Pascal VOC 2007 dataset. * indicates the result is reproduced by using the open-source code, and - denotes the corresponding result is not provided.}
  \label{tab:voc-results}
\end{table}

\noindent{\textbf{Performance on Pascal VOC 2007. }} Pascal VOC 2007 is the most widely used dataset for evaluating multi-label image recognition. Here, we also present the performance comparisons on this dataset in Table \ref{tab:voc-results}. Although current algorithms can achieve quite well performance on this dataset, our SRDL framework can still achieve about the same performance as state-of-the-art works. As shown, our framework achieves the best mAP of 94.6\%, slightly better than the second best mAP of 94.4\%.

\begin{table}[!h]
  \centering
  \small
  \begin{tabular}{|c|c|cc|c|c|}
  \hline
  Line No. & Methods & CAR & OE & MS-COCO & VOC 2007\\
  \hline
  1 & Baseline &  &  & 77.1 & 91.8 \\
  \hline
  2 & \multirow{2}{*}{Ours} & \checkmark &  & 80.3 ({\color{red}$\uparrow$ 3.2}) & 92.4 ({\color{red}$\uparrow$ 0.6}) \\
  3 & ~ & \checkmark & \checkmark & 82.9 ({\color{red}$\uparrow$ 5.8}) & 94.6 ({\color{red}$\uparrow$ 2.8}) \\
  \hline
  \end{tabular}
  \vspace{2pt}
  \caption{Ablative study of CAR and OE module on the MS-COCO and Pascal VOC 2007 datasets in mAP (\%).}
  \label{tab:ablation-result}
\end{table}

\subsection{Ablation Study}
The proposed SRDL framework builds on the ResNet-101 \cite{He2016ResNet}, thus we compare with this baseline to analyze the contribution of the proposed framework. As shown in Table \ref{tab:ablation-result}, our framework significantly outperforms the baseline on MS-COCO and Pascal VOC 2007, improving the mAP from 77.1\% and 91.8\% to 82.9\% a 94.6\%, respectively. 

The foregoing comparisons verify the effectiveness of the proposed SRDL framework as a whole. Actually, the SRDL framework contains two critical modules, i.e., category-specific attentional regions (CAR) module and object erasing (OE) module. In the following, we further conduct ablative experiments to analyze the actual contribution of each module.

\begin{table}[!h]
  \centering
  \small
  \begin{tabular}{|c|c|c|c|}
  \hline
  Line No. & Methods & MS-COCO & VOC 2007 \\
  \hline
  1 & Baseline & 77.1 & 91.8 \\
  \hline
  2 & SD & 80.1 & 92.2 \\
  3 & SAM & 79.9 & 92.0 \\
  \hline
  4 & Ours CAR w/o GCN & 78.4 & 92.0 \\
  5 & Ours CAR w/o CA & 79.7 & 92.1 \\
  6 & Ours CAR w/o SA & 79.3 & 92.0 \\
  7 & Ours CAR & \textbf{80.3} & \textbf{92.4} \\
  \hline
  \end{tabular}
  \vspace{2pt}
  \caption{Comparison of the baseline,  semantic decoupling module (SD), semantic attention module (SAM), our framework merely using CAR without GCN (Ours CAR w/o GCN), our framework merely using CAR without channel-wise attention (Ours CAR w/o CA), our framework merely using CAR without spatial-wise attention (Ours CAR w/o SA) and our framework merely using CAR (Ours) on the MS-COCO and Pascal VOC 2007 datasets in mAP (\%).}
  \label{tab:semantic-learning-result}
\end{table}

\begin{figure}[!h]
   \centering
   \includegraphics[width=0.95\linewidth]{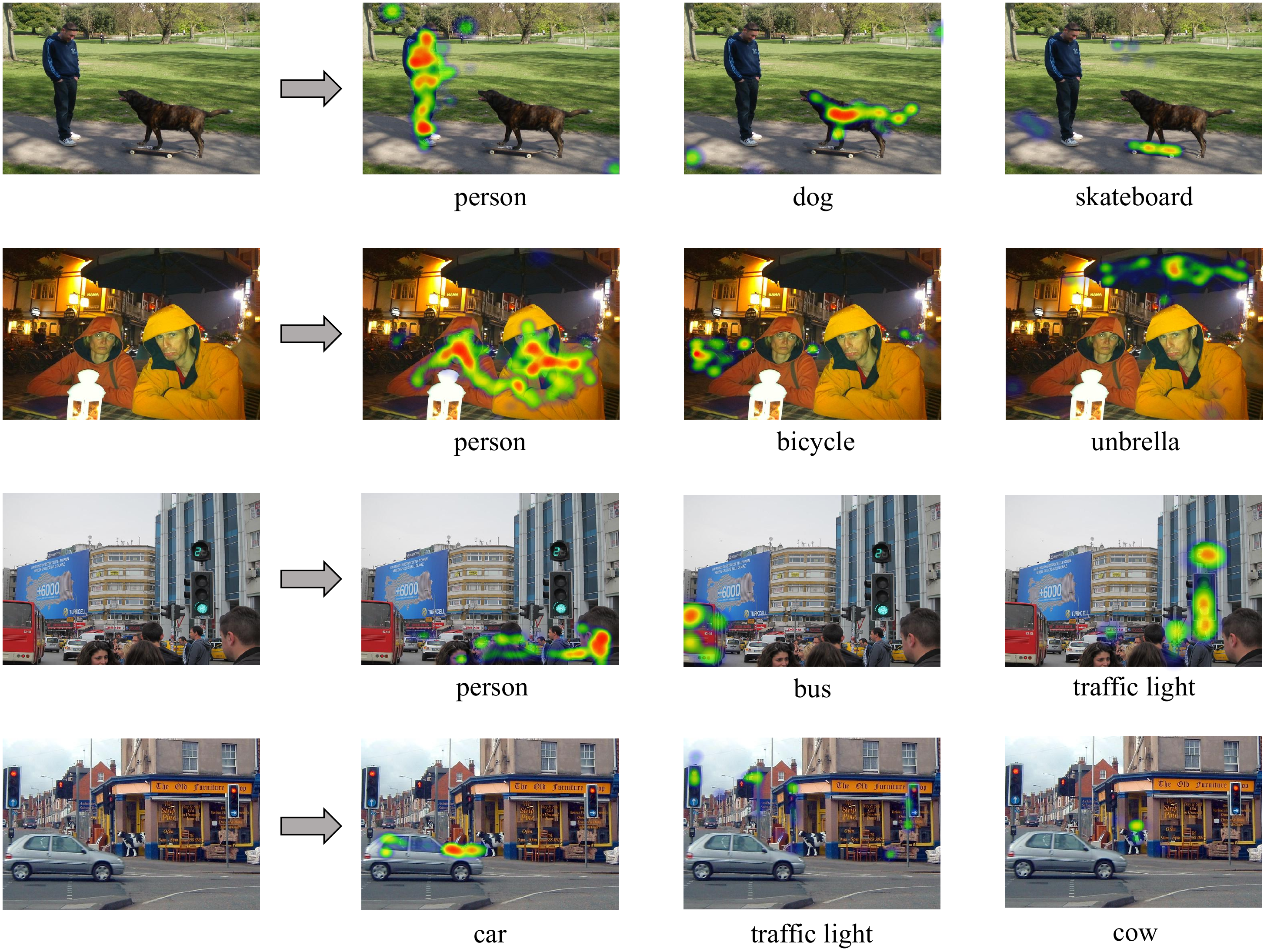}
   \caption{Several examples of input images (left) and 
   spatial-wise attention maps corresponding to categories with top 3 highest confidences (right).}
   \label{fig:visual}
\end{figure}

\noindent{\textbf{Analysis of CAR module. }} To explore the actual contribution of the CAR module, we conduct experiments that merely use this module and compare it with the baseline on MS-COCO and Pascal VOC 2007 datasets. As presented in Table \ref{tab:ablation-result} (line 2), it obtains the mAP of 80.3\% and 92.4\% on MS-COCO and Psacal VOC 2007, with an improvement of 3.2\% and 0.6\%, respectively.  Such improvement shows that the CAR module can guide model to learn category-specific semantic representation by generating channel/spatial-wise attention maps for each category. Meanwhile, it is worth noting that the CAR module not only localizes semantic-aware regions in spatial dimension but also models inter-dependencies between channels.

As a primary component of our proposed framework, we present detailed and comprehensive ablation results for the CAR module. As shown in Table \ref{tab:semantic-learning-result}, our CAR module outperforms current category-specific semantic learning modules (i.e., SD \cite{Chen2019SSGRL} and SAM \cite{Ye2020ADD-GCN}), which demonstrates using both spatial-wise and channel-wise attention can help to learn category-specific semantic representation better. Besides, we also conduct more experiments for in-depth understanding. Specifically, we add the CAR module without GCN to the baseline ResNet-101, namely ``Ours CAR w/o GCN”, and compare it with our framework merely using CAR (namely ``Ours CAR") to verify the contribution of GCN. The experiment results in Table \ref{tab:semantic-learning-result} shows that removing the GCN in the CAR module degrades the average mAP from 80.3\% to 78.4\% on the MS-COCO dataset. Here, we also evaluate respectively the contribution of the spatial/channel-wise attention in the CAR module, as show in Table \ref{tab:semantic-learning-result}. It is observed that the average mAP decreases from 80.3\% to 79.7\% on the MS-COCO dataset while removing the channel-wise attention and the average mAP decreases from 80.3\% to 78.9\% on the MS-COCO dataset while removing the spatial-wise attention.

As discussed above, the CAR module can generate precise spatial-wise attention maps for each category to guide model to focus on semantic-aware regions to learn more discriminative category-specific semantic representation. Here, we further visualize some examples in Figure \ref{fig:visual}. In each row, we present the input image, the spatial-wise attention maps corresponding to categories with the top 3 highest confidences. As can be observed, the third example has objects of ``person", ``bus" and ``traffic light", the CAR module not only precisely highlights the obvious corresponding semantic-aware regions of these categories, but also highlights the easily overlooked semantic-aware regions of small object of ``traffic light". It intuitively shows that the CAR module can well localize the semantic-aware regions. 

\begin{figure}[!t]
   \centering
   \includegraphics[width=0.45\linewidth]{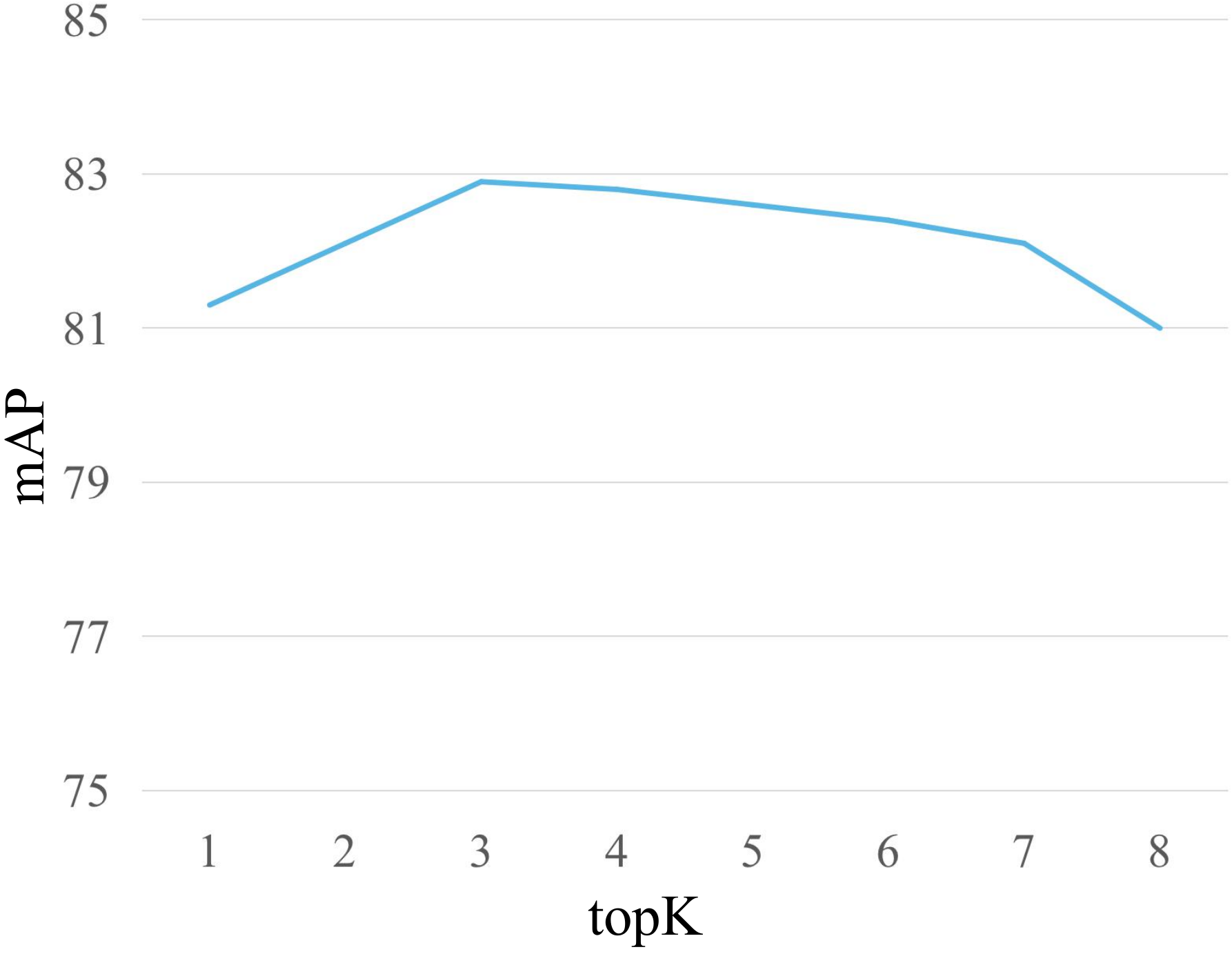}~~~
   \includegraphics[width=0.45\linewidth]{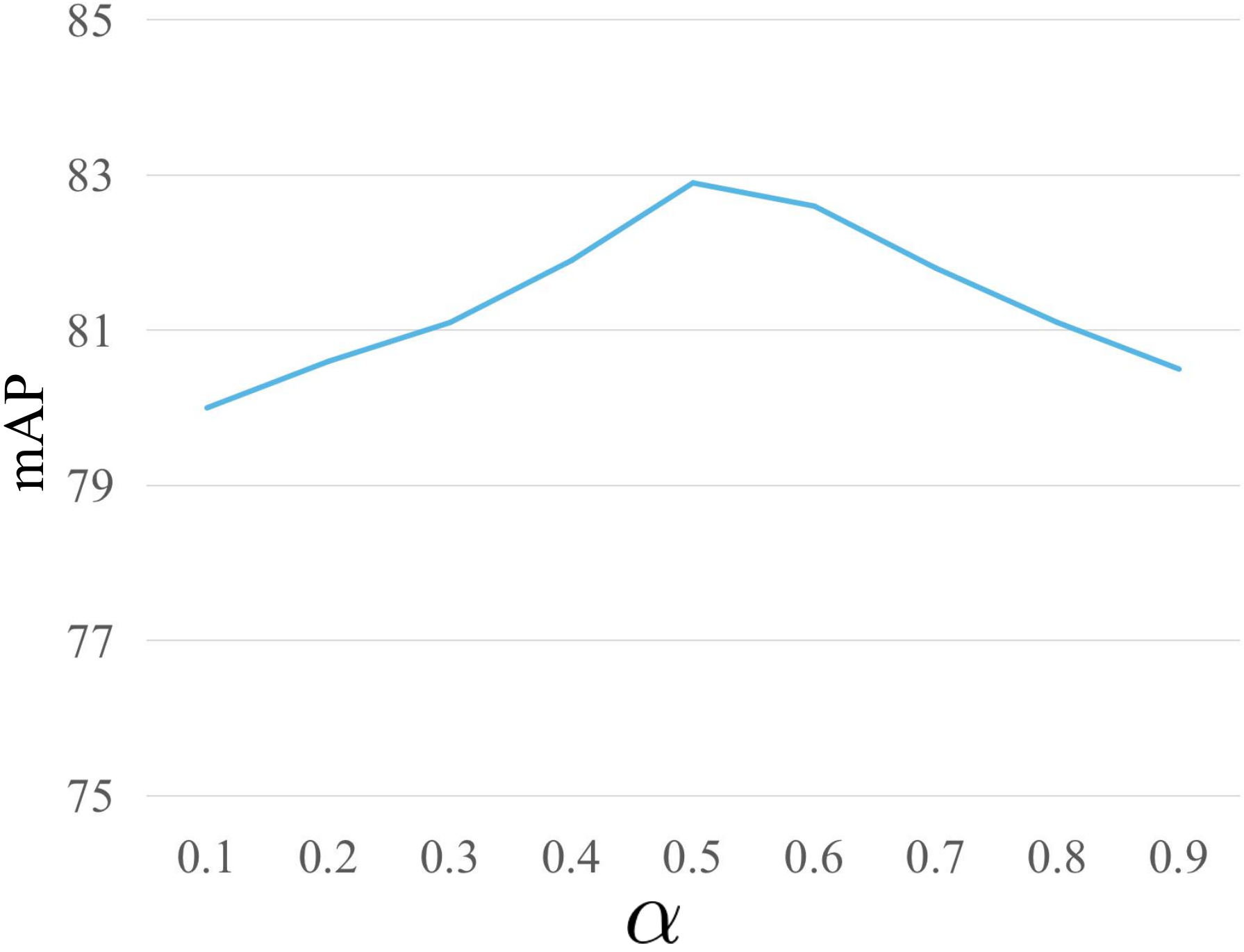}
   \caption{The performance of our framework with different values of \textit{topK} (left) and $\alpha$ (right).}
   \label{fig:ablation}
\end{figure}

\noindent{\textbf{Analysis of OE module. }} For figuring out the actual contribution of the OE module, we compare the performance with and without this module. As shown in Table \ref{tab:ablation-result} (line 3 vs. line 2), we can see that our framework with the OE module performs better than our framework without the OE module, improving the mAP from 80.3\% and 92.4\% to 82.9\% and 94.6\% on MS-COCO and Psacal VOC 2007. One of the most likely reasons is that the network may learn a more robust and contextualized representation for each category and capture semantic dependency among all categories by adaptively erasing semantic-aware regions during training.

\begin{figure}[!h]
   \centering
   \includegraphics[width=0.9\linewidth]{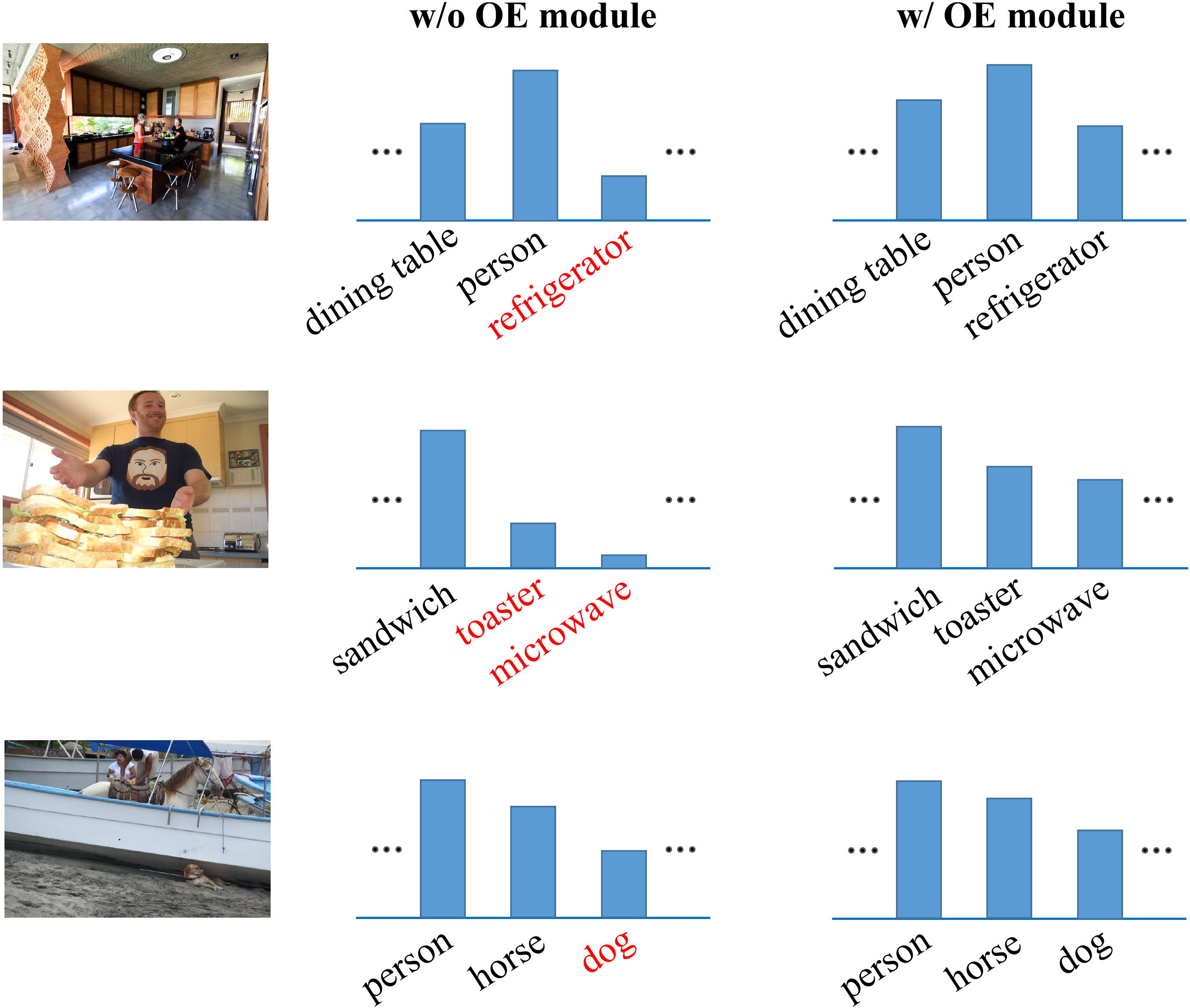}
   \caption{Several examples of input images (left), predicted scores without OE module (middle), and predicted scores with OE module (right). Uncorrected labels are highlighted in red.}
   \label{fig:visual1}
\end{figure}

Except the above discussion, we also conduct  experiments to explore the impact of \textit{topK}. Specifically, we fix $\alpha$ to 0.5 and choose the value \textit{topK} from a given set $\{1, 2, 3, 4, 5, 6, 7, 8\}$. As presented in Figure \ref{fig:ablation}, the mAP performance shows an upward trend with the value of \textit{topK} gradually being increased from \textit{topK}=1, and the performance tends to be stable when \textit{topK} is set to 3. However, the performance significantly drops when \textit{topK} is set to high value (e.g., 7 or 8). One likely reason is that erasing too many semantic-aware regions greatly improves the difficulty of recognizing objects which may disturb the network training and degrade model performance.

Furthermore, we also perform experiments that fix \textit{topK} to 3 and vary the minimum $\alpha$ from 0.1 to 0.9 to explore the sensitivity of the threshold $\alpha$. As shown in Figure \ref{fig:ablation}, the performance drops when $\alpha$ is either too small or too large, because a too small $\alpha$ results in erasing a too large semantic-aware region which significantly degrades model performance, and a too large $\alpha$ results in erasing a too small semantic-aware region which makes semantic dependency learning ineffective.

\section{Conclusion}
Learning more discriminative and robust semantic representation for each category and capturing semantic dependency among all categories are the key of solving the multi-label image recognition. In this work, to achieve these purposes, we propose a novel and effective semantic representation and dependency learning (SRDL) framework which consists of two crucial modules, i.e., a category-specific attentional regions (CAR) module that guides model to learn more discriminative semantic represenation for each category $c$ by generating channel/spatial-wise attention maps, and an object erasing (OE) module that guides model to capture semantic dependency among categories by adaptively erasing semantic-aware regions. Extensive experiments on two benchmarks including MS-COCO and Pascal VOC 2007 demonstrate the effectiveness of the proposed framework over existing leading methods.

\section*{Acknowledgement}

This work is supported in part by National Key R\&D Program of China under Grant No. 2021ZD0111601, National Natural Science Foundation of China (NSFC) under Grant No. 61876045, 62272494, U1811463 and 61836012, Guangdong Basic and Applied Basic Research Foundation under Grant No. 2017A030312006, and Key Laboratory open fund under Grant No. WDZC20215250118.

\bibliography{mybibfile}

\end{document}